# Forecasting Foreign Exchange Market Prices Using Technical Indicators with Deep Learning and Attention Mechanism

Sahabeh Saadati
Department of Computer Engineering, North Tehran Branch, Islamic Azad University,
Tehran, Iran
S.saadati@iau-tnb.ac.ir

Mohammad Manthouri
Department of Electrical and Electronic Engineering
Shahed University,
Tehran, Iran
mmanthouri@shahed.ac.ir

*Abstract*— **Accurate prediction of price behavior in the foreign exchange market is crucial. This paper proposes a novel approach that leverages technical indicators and deep neural networks. The proposed architecture consists of a Long Short-Term Memory (LSTM) and Convolutional Neural Network (CNN), and attention mechanism. Initially, trend and oscillation technical indicators are employed to extract statistical features from Forex currency pair data, providing insights into price trends, market volatility, relative price strength, and overbought and oversold conditions. Subsequently, the LSTM and CNN networks are utilized in parallel to predict future price movements, leveraging the strengths of both recurrent and convolutional architectures. The LSTM network captures long-term dependencies and temporal patterns in the data, while the CNN network extracts local patterns. The outputs of the parallel LSTM and CNN networks are then fed into an attention mechanism, which learns to weigh the importance of each feature and temporal dependency, generating a context-aware representation of the input data. The attention-weighted output is then used to predict future price movements, enabling the model to focus on the most relevant features and temporal dependencies. Through a comprehensive evaluation of the proposed approach on multiple Forex currency pairs, we demonstrate its effectiveness in predicting price behavior and outperforming benchmark models.**

I. INTRODUCTION

The foreign exchange (Forex) market, being the largest financial market globally, with a daily trading volume exceeding $5 trillion, presents a multitude of opportunities for traders and investors. This market facilitates the simultaneous trading of various countries' currencies against one another. In the Forex market, currency pairs are formed, which are then traded against each other [1]. The high liquidity and volatility of the Forex market make it an attractive platform for traders seeking to capitalize on exchange rate fluctuations. Furthermore, the market's operational hours, spanning across different time zones, enable traders to trade at any time. As a result, the Forex market has become a popular destination for individuals and institutions alike, seeking to diversify their investment portfolios and manage risk [2]. Therefore, accurate prediction of exchange rate movements is crucial for traders and investors to make informed investment decisions, minimize potential losses, and maximize returns. Effective forecasting enables market participants to stay ahead of market trends, adjust their strategies accordingly, and ultimately gain a competitive edge in the highly volatile Forex market [3]. Silicon Valley's technological innovations and venture capital have played a crucial role in strengthening global financial markets by fostering entrepreneurship, driving economic growth, and increasing access to capital for startups and small businesses [4]. [5] proposes an innovative approach to agricultural education utilizes immersive technologies, such as augmented reality and virtual reality, to gamify learning and create engaging, interactive experiences that enhance student knowledge and skills. The Grace Platform is a work-in-progress initiative that integrates gamified augmented reality and virtual reality to revolutionize agriculture education, aiming to enhance pedagogy and provide immersive, interactive learning experiences for students [6].

Technical indicators play a vital role in Forex trading, analysis, and prediction, as they provide valuable insights into market trends, patterns, and sentiment, enabling traders to make more informed investment decisions. For example, Moving Averages (MA), such as the 50-day and 200-day averages, help traders identify trends and predict potential reversals. The Relative Strength Index (RSI) and is popular indicator used to detect overbought and oversold conditions, allowing traders to adjust their strategies accordingly. Bollinger Bands, which consist of a moving average and two standard deviations, provide a visual representation of volatility and can be used to identify potential breakouts and trend changes [7]. These technical indicators effectively operate as a feature extractor, distilling the complexities of the market into a set of meaningful and relevant features. These extracted features can then be leveraged by predictive models to identify patterns, trends, and relationships that might otherwise remain hidden in Forex price data [8].

Long Short-Term Memory (LSTM) networks, a type of Recurrent Neural Network (RNN), have emerged as a powerful tool for predicting time series data. By leveraging their ability to learn long-term dependencies and patterns in sequential data, LSTMs can effectively capture the complex dynamics underlying time series data [9]. Unlike traditional RNNs,

LSTMs are capable of mitigating the vanishing gradient problem, allowing them to learn and remember information over long periods of time [10]. This makes them particularly well-suited for predicting time series data, such as Forex prices, where patterns and trends can emerge over extended periods. As a category of deep learning models, LSTMs are able to automatically extract relevant features from raw data, eliminating the need for manual feature engineering. By training an LSTM network on historical time series data, it is possible to generate accurate predictions of future values, enabling informed decision-making and strategic planning [11]. One-dimensional Convolutional Neural Networks (CNNs 1D) are highly effective for time series prediction due to their ability to capture local patterns and temporal dependencies within sequential data. By applying convolutional filters along the time axis, CNN 1D can automatically learn important features such as trends, seasonality, and irregular patterns from raw time series data [12]. [13] leverages deep learning techniques to predict financial market sequences, aiming to provide valuable insights for policymakers to enhance economic policies and make informed decisions.

Attention mechanisms have revolutionized the field of time series analysis and Natural Language Processing (NLP) by enabling models to selectively focus on the most relevant segments of data [14]. In time series analysis, attention mechanisms allow models to weigh the importance of different time steps, identifying the most influential patterns and trends that drive future values [15]. This is particularly useful in cryptocurrency price prediction, where attention can be directed towards specific time periods or events that have a significant impact on future prices [16]. Attention mechanisms have also proven effective in forex prediction by enabling models to prioritize critical time steps and economic indicators, thus enhancing the accuracy of exchange rate forecasts. By focusing on influential market events and temporal patterns, attention-based models can better capture the complexities of forex markets, providing more reliable predictions for traders and analysts.

This paper proposes a novel hybrid approach that combines technical indicators with parallel LSTM and CNN networks and an attention mechanism, achieving improved accuracy in predicting Forex price behavior and outperforming benchmark models. Here are the contributions of the paper in three bullet points:

• **Feature Extraction:** The paper employs trend and oscillation technical indicators to extract statistical features from Forex currency pair data, providing insights into price trends, market volatility, relative price strength, and overbought and oversold conditions.

• **Parallel CNN-LSTM Architecture**: The paper utilizes a novel parallel architecture that combines the strengths of CNN and LSTM networks to predict future price movements.

• **Attention-based Output**: The outputs of the parallel CNN and LSTM networks are fed into an attention mechanism, which learns to weigh the importance of each feature and temporal dependency, generating a context-aware representation of the input data and enabling the model to focus on the most relevant features and temporal dependencies.

The proposed method is evaluated on three major Forex currency pairs, CHFUSD, EURUSD, and GBPUSD, and demonstrates its effectiveness in predicting price behavior and outperforming benchmark models.

This paper is structured as follows: Section II introduces the Forex currency dataset and provides a literature review. Section III delves into the methodology, which involves the development of CNN, LSTM, and hybrid LSTM-CNN-Attention models. The simulation results are presented in Section IV, where the training and evaluation of the proposed methods are discussed in the context of hourly CHFUSD, EURUSD, and GBPUSD datasets. Finally, the paper concludes in Section V by summarizing the key findings and contributions of this research.

## II. CURRENCY MARKET FORECASTING

This section highlights the one-hour time frame as a popular choice among forex traders and explores key currency pairs commonly used in this timeframe. Also, the section reviews recent studies on applying deep learning techniques, such as CNNs, LSTMs, and attention mechanisms, to improve forex predictions and highlights their strengths, limitations, and potential to revolutionize forex market analysis and prediction [17].

### A. Forex Dataset

The one-hour time frame offers a unique perspective on market dynamics, allowing traders and analysts to capture short-term fluctuations and trends in currency prices. With a one-hour time frame, market participants can identify and respond to rapid changes in market sentiment, news-driven events, and technical indicators, making it an ideal time frame for intraday trading and scalping strategies [3]. By focusing on this shorter time frame, traders can capitalize on fleeting opportunities and minimize exposure to overnight risks, making it an attractive option for those seeking to maximize returns in a fast-paced and volatile market environment.

The three major currency pairs, CHFUSD, EURUSD, and GBPUSD, are widely traded and closely watched in the foreign exchange market. With a one-hour time frame, these currency pairs offer a unique opportunity to analyze and forecast short-term price movements. The CHFUSD pair, representing the Swiss franc against the US dollar, is known for its stability and low volatility. The EURUSD pair, representing the euro against the US dollar, is one of the most liquid and widely traded currency pairs in the world. The GBPUSD pair, representing the British pound against the US dollar, is heavily influenced by economic indicators and geopolitical events [18]. By examining these three currency pairs with a one-hour time frame, traders and analysts can gain valuable insights into market trends and make informed trading decisions.

### B. Literature Survey

The application of CNN in Forex prediction has shown promising results in identifying patterns and trends in currency exchange rates, enabling more accurate and informed trading decisions. [19] proposes a CNN approach to predict forex trends, leveraging the strengths of CNNs in image processing to analyze and forecast currency market patterns. [20] proposes a novel approach to cryptocurrency price forecasting using CNNs with weighted and attentive memory channels, enabling the model to capture complex patterns and trends in cryptocurrency markets. [21] presents a novel CNN-LSTM-based approach for medium to long-term copper price prediction, incorporating multiple influencing factors to improve the accuracy and reliability of price forecasts in the copper market. RoboMan is a cutting-edge, adult-sized humanoid robot designed to facilitate research on humanoids, boasting enhanced performance, inherent stability, and advanced two-stage balance control [22].

The application of LSTM networks in Forex prediction has demonstrated effectiveness in capturing temporal dependencies and nonlinear relationships in currency exchange rates, leading to improved forecasting accuracy and trading performance. [23] presents a novel approach to stock price forecasting by combining XGBoost and LSTM models, demonstrating improved accuracy and promising results for stock market predictions. [24] proposes a novel approach to forex price prediction by combining wavelet denoising with an attention-based RNN-ARIMA model, achieving improved accuracy and robustness in forecasting currency exchange rates. [25] optimizes LSTM models for EUR/USD prediction, evaluating the trade-off between model performance and energy consumption, providing insights for energy-efficient forecasting in the foreign exchange market. [26] introduces FLF-LSTM, a novel prediction system that utilizes a custom Forex Loss Function to improve the accuracy and robustness of LSTM-based forex price predictions. [27] investigates the relationship between cross-model neuronal correlations and model performance, shedding light on the neural mechanisms underlying predictive power and generalizability in deep learning models.

The incorporation of attention mechanisms in Forex prediction models has enabled the selective focus on relevant market indicators and time steps, leading to more accurate and interpretable predictions of currency exchange rates. [28] introduces a novel variant of LSTM-based stock prediction method that incorporates an attention mechanism, enabling the model to focus on relevant input features and improve the accuracy of stock price forecasts. [29] This paper proposes an attention-based CNN-LSTM model for high-frequency multiple cryptocurrency trend prediction, leveraging the strengths of both CNNs and LSTMs to capture complex patterns and trends in cryptocurrency markets. [9] presents a novel approach to cryptocurrency price prediction by combining transformer neural networks with technical indicators, enabling the model to capture long-range dependencies and improve the accuracy of price forecasts. [30] proposes a novel stock series prediction model that integrates variational mode decomposition with a dual-channel attention network, allowing the model to extract complex patterns and trends from stock data and improve forecasting accuracy.

III. METHODOLOGY

The third section of the paper is divided into two primary components. The first part focuses on the preprocessing of technical indicators, while the second part provides a detailed explanation of the proposed hybrid model, which is specifically designed for time series forecasting of forex currencies.

A. Technical Indicator

Technical indicators are widely used in finance to analyze and predict market trends. These indicators can be leveraged as feature extractors in machine learning and deep learning models to provide valuable insights into market behavior. In this context, technical indicators can be seen as a way to transform raw market data into meaningful features that can be used to train predictive models. The formula for a Simple Moving Average is:

$$SMA = \frac{1}{n}\sum_{i=1}^{n} P_i \qquad (1)$$

where $\sum_{i=1}^{n} P_i$ is the sum of the closing prices of the last $n$ periods. The RSI (Relative Strength Index) calculates the ratio of average gains to average losses over a certain period of time to identify when an instrument is overbought or oversold. The formula is:

$$RSI = 100 - 100 / (1 + RS) \qquad (2)$$

where RS is the average gain divided by the average loss, both calculated over a specified number of periods. Bollinger Bands are a technical indicator consisting of a central band accompanied by two outer bands.

$$Upper - Lower\ Band = SMA(n) \pm k \times std(n) \qquad (3)$$

$$Middle\ Band = SMA(n) \qquad (4)$$

This configuration enables the bands to adapt to changes in volatility, thereby facilitating the identification of overbought and oversold market conditions.

B. Hybrid LSTM-CNN-Attention Model

The proposed model is a sequential neural network designed specifically for predicting time series data. It's composed of multiple interconnected layers that work together to identify meaningful patterns in sequential data. The first layer, which is shaped to match the input time series, serves as the input layer. The LSTM layer, a key component of the model, can be mathematically represented by a set of equations.

$$f_t = \sigma(W_f.[h_{t-1}, x_t] + b_f) \qquad (5)$$
$$i_t = \sigma(W_i.[h_{t-1}, x_t] + b_i) \qquad (6)$$
$$\tilde{C}_t = tanh(W_C.[h_{t-1}, x_t] + b_C) \qquad (7)$$
$$C_t = f_t \times C_{t-1} + i_t * \tilde{C}_t \qquad (8)$$
$$o_t = \sigma(W_o.[h_{t-1}, x_t] + b_o) \qquad (9)$$
$$h_t = o_t \times tanh(C_t) \qquad (10)$$

The LSTM layer is defined by three gates: the forget gate ($f_t$), input gate ($i_t$), and output gate ($o_t$). These gates control the flow of information into and out of the cell state ($C_t$) and hidden state ($h_t$). The input at time step $t$ is denoted as $x_t$. The sigmoid function ($\sigma$) is used to introduce non-linearity into the gates. The weight matrices ($W$) and bias vectors ($b$) are learnable parameters that are used to compute the gates and states.

The second layer is a 1D Convolutional Neural Network (CNN1D) designed to extract local patterns and features from the time series data. By parallelizing the CNN1D layer with the LSTM layer, the model can leverage the strengths of both architectures to improve predictive performance. The CNN1D layer applies a set of filters to the input data, scanning the time series sequence to identify local patterns and trends. This allows the model to capture both short-term and long-term dependencies in the data, providing a more comprehensive understanding of the time series dynamics.

$$Y[t] = \sum_{k=1}^{K}(X[t-k+1:t].W_c[:,k-1]) + b_c \quad (11)$$

Where $Y[t]$ is the output at time step $t$, $X[t-k+1:t]$ is the input sequence from time step $t-k+1$ to $t$, $W_c[:,k-1]$ is the weight matrix slice from column $k-1$ to the end, $b$ is the bias term and $k$ is the kernel size.

The output of the LSTM layer and the CNN1D layer are concatenated to form a single output vector, denoted as Z[t].

$$Z[t] = Concat(h_t, Y[t]) \quad (12)$$

The third layer introduces an attention mechanism, which is applied to the concatenated output $Z[t]$ from the previous layers. Specifically, $Z[t]$ is the result of concatenating the hidden state $h_t$ from the LSTM layer and the output $Y[t]$ from the CNN1D layer. The attention mechanism allows the model to selectively focus on the most relevant parts of the concatenated output, weighing their importance to inform the final prediction. By applying attention to $Z[t]$, the model can dynamically adjust its emphasis on different segments of the time series data, leading to more accurate and informed predictions. The attention mechanism is defined by the following equation

$$\alpha = softmax(W_a.Z[t] + b_a) \quad (13)$$
$$c_t = \sum((\alpha) \times Z[t]) \quad (14)$$

Here, $W_a$ and $b_a$ are learnable weights and bias terms, respectively, for the attention mechanism. The attention weights $\alpha$ are computed using the softmax function, and the context vector $c_t$ is computed as a weighted sum of the concatenated output vector $Z[t]$. The attention weights $\alpha$ learn to assign higher importance to certain time steps or features in the input sequence, which helps the model to make more accurate predictions. The model concludes with a straightforward dense layer, which is defined as follows

$$y_d = softmax(W_d.c_t + b_d) \quad (15)$$

where $y_d$ is the prediction output vector, and $W_d$ and $b_d$ are learnable parameters.

## IV. SIMULATION RESULTS

The three major currency pairs are CHFUSD, EURUSD, and GBPUSD, with a one-hour time frame. The training data includes data from the beginning of 2020 to the beginning of 2024, with 20% allocated for training and 80% reserved for testing. In this section, we conduct an empirical investigation of the performance of these three major Forex currency pairs using the CNN1D, LSTM, and Hybrid LSTM-CNN-Attention models. This analysis aims to validate and substantiate the claims made about the effectiveness of the proposed model in predicting Forex prices. Figures 1-3 illustrate the predicted outcomes of the CNN1D, LSTM, and Hybrid LSTM-CNN-Attention models, providing a visual representation of their performance in forecasting the CHFUSD, EURUSD, and GBPUSD currency pairs.

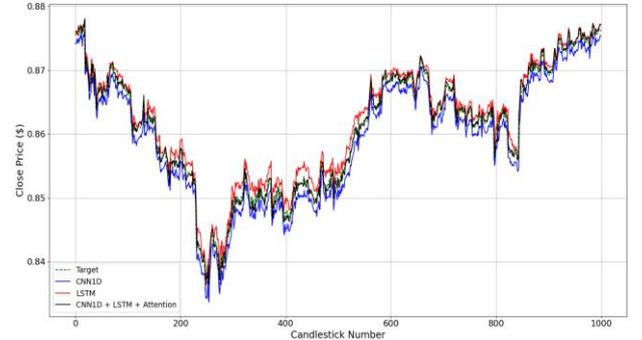

Figure 1: Predicted vs. Actual Values for CHFUSD Currency Pair using CNN1D, LSTM, and Hybrid LSTM-CNN-Attention Models

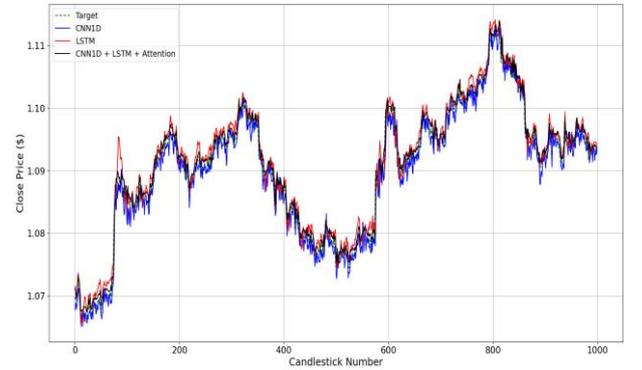

Figure 2: Predicted vs. Actual Values for EURUSD Currency Pair using CNN1D, LSTM, and Hybrid LSTM-CNN-Attention Models

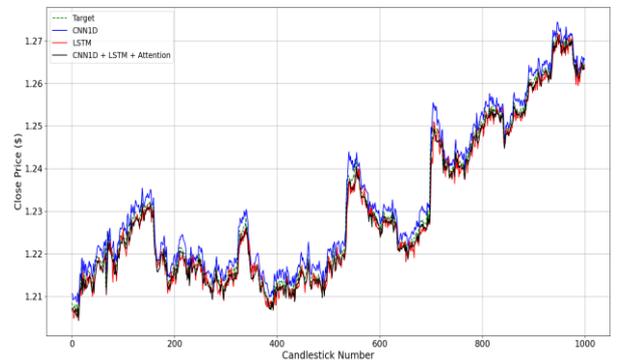

Figure 3: Predicted vs. Actual Values for GBPUSD Currency Pair using CNN1D, LSTM, and Hybrid LSTM-CNN-Attention Models

Tables 1-3 present the performance metrics of the CNN1D, LSTM, and Hybrid LSTM-CNN-Attention models for the CHFUSD, EURUSD, and GBPUSD currency pairs, including Mean Squared Error (MSE), Root Mean Squared Error

(RMSE), and R-Square values. The performance metrics used are:

- MSE (Mean Squared Error): a measure of the average squared difference between predicted and actual values.
- RMSE (Root Mean Squared Error): the square root of MSE, which is a measure of the average distance between predicted and actual values.
- R-Square (Coefficient of Determination): a measure of how well the model explains the variance in the data, with values close to 1 indicating a good fit.

Table 1. Performance prediction of CHFUSD using CNN1D, LSTM, and CNN1D- LSTM-Attention neural networks

|  | MSE | RMSE | R − Square |
|---|---|---|---|
| CNN1D | 2.9e-06 | 0.00171 | 0.99821 |
| LSTM | 2.6e-06 | 0.00164 | 0.99844 |
| CNN1D- LSTM-Attention | 1.7e-06 | 0.00134 | 0.99941 |

Table 2. Performance prediction of EURUSD using CNN1D, LSTM, and CNN1D- LSTM-Attention neural networks

|  | MSE | RMSE | R − Square |
|---|---|---|---|
| CNN1D | 3.7e-06 | 0.00194 | 0.99842 |
| LSTM | 2.0e-06 | 0.00144 | 0.99907 |
| CNN1D- LSTM-Attention | 1.4e-06 | 0.00121 | 0.9997 |

Table 3. Performance prediction of GBPUSD using CNN1D, LSTM, and CNN1D- LSTM-Attention neural networks

|  | MSE | RMSE | R − Square |
|---|---|---|---|
| CNN1D | 4.1e-06 | 0.00204 | 0.99812 |
| LSTM | 3.6e-06 | 0.00191 | 0.99890 |
| CNN1D- LSTM-Attention | 1.9e-06 | 0.0014 | 0.99945 |

The figures illustrate the superior performance of the CNN1D-LSTM-Attention model, which is evident from the plots. The CNN1D-LSTM-Attention model demonstrates a closer fit to the actual exchange rates, highlighting its ability to accurately forecast currency prices.

From the tables, we can observe that in all three currency pairs, the CNN1D-LSTM-Attention model outperforms the other two models in terms of MSE, RMSE, and R-Square. This suggests that the attention mechanism in the CNN1D-LSTM-Attention model is effective in capturing the complex patterns in the exchange rate data. The LSTM model generally performs better than the CNN1D model, indicating that the recurrent structure of the LSTM model is more suitable for modeling the temporal dependencies in the exchange rate data.

Overall, the results confirm that the CNN1D-LSTM-Attention model is the most effective in predicting exchange rates among the three models evaluated.

## V. CONCLUSION

This paper proposes a novel hybrid approach that combines technical indicators with parallel LSTM and CNN networks and an attention mechanism, achieving improved accuracy in predicting Forex price behavior. The proposed method extracts statistical features from Forex currency pair data using trend and oscillation technical indicators, leverages a parallel CNN-LSTM architecture to predict future price movements, and utilizes an attention mechanism to generate a context-aware representation of the input data. The results demonstrate the effectiveness of the proposed method in predicting price behavior and outperforming benchmark models across three major Forex currency pairs, CHFUSD, EURUSD, and GBPUSD. The superior performance of the CNN1D-LSTM-Attention model is evident from the plots, which show a closer fit to the actual exchange rates. The attention mechanism is effective in capturing complex patterns in the exchange rate data, and the recurrent structure of the LSTM model is more suitable for modeling temporal dependencies. Overall, the proposed method offers a promising approach for Forex price prediction and has the potential to be applied in real-world trading and investment scenarios.

Future work directions include exploring the application of the proposed hybrid approach to other financial markets, such as stocks and commodities, to further validate its generalizability and effectiveness. Additionally, incorporating other types of data, such as news articles, social media sentiment, and macroeconomic indicators, into the model could provide a more comprehensive view of the market and potentially improve prediction accuracy. Furthermore, experimenting with different attention mechanisms and neural network architectures could lead to further performance enhancements. Finally, developing a real-time trading system that integrates the proposed model with risk management and portfolio optimization techniques could enable the creation of a fully automated trading platform.